\documentclass{article}

\usepackage[nonatbib, final]{neurips_2020}

\usepackage[utf8]{inputenc} 
\usepackage[T1]{fontenc}    
\usepackage{hyperref}       
\usepackage{url}            
\usepackage{booktabs}       
\usepackage{nicefrac}       
\usepackage{microtype}      
\usepackage{textcomp}

\usepackage{amsmath,amsfonts,bm}

\def\eqref#1{equation~\ref{#1}}

\def\1{\bm{1}}

\def\vb{{\bm{b}}}

\def\vf{{\bm{f}}}

\def\vk{{\bm{k}}}

\def\vp{{\bm{p}}}

\def\vr{{\bm{r}}}

\def\vx{{\bm{x}}}
\def\vy{{\bm{y}}}

\def\mK{{\bm{K}}}

\def\mM{{\bm{M}}}

\def\mP{{\bm{P}}}

\def\mW{{\bm{W}}}
\def\mX{{\bm{X}}}

\DeclareMathAlphabet{\mathsfit}{\encodingdefault}{\sfdefault}{m}{sl}
\SetMathAlphabet{\mathsfit}{bold}{\encodingdefault}{\sfdefault}{bx}{n}

\newcommand{\R}{\mathbb{R}}

\usepackage{amssymb, amsmath,amsfonts,amsthm,bm} 
\usepackage{mathtools}
\usepackage{multirow}
\usepackage{caption}
\usepackage{booktabs}
\usepackage{float}
\usepackage{changepage}
\usepackage{array}
\usepackage{rotating}
\usepackage{makecell}

\newcolumntype{P}[1]{>{\centering\arraybackslash}p{#1}}

\title{Permutation Matters: Anisotropic Convolutional Layer for Learning on Point Clouds}

\author{%
  Zhongpai~Gao, Guangtao~Zhai, Junchi~Yan, and Xiaokang~Yang\\
  Artificial Intelligence Institute, Shanghai, China\\
  Shanghai Jiao Tong University\\
  Shanghai, 200240, China \\
  \texttt{\{gaozhongpai, zhaiguangtao, yanjunchi, xkyang\}@sjtu.edu.cn}
}

\begin{document}

\maketitle

\begin{abstract}
   It has witnessed a growing demand for efficient representation learning on point clouds in many 3D computer vision applications. Behind the success story of convolutional neural networks (CNNs) is that the data (e.g., images) are Euclidean structured. However, point clouds are irregular and unordered. Various point neural networks have been developed with isotropic filters or using weighting matrices to overcome the structure inconsistency on point clouds. However, isotropic filters or weighting matrices limit the representation power. In this paper, we propose a permutable anisotropic convolutional operation (PAI-Conv) that calculates soft-permutation matrices for each point using dot-product attention according to a set of evenly distributed kernel points on a sphere's surface and performs shared anisotropic filters. In fact, dot product with kernel points is by analogy with the dot-product with keys in Transformer as widely used in natural language processing (NLP). From this perspective, PAI-Conv can be regarded as the transformer for point clouds, which is physically meaningful and is robust to cooperate with the efficient random point sampling method. Comprehensive experiments on point clouds demonstrate that PAI-Conv produces competitive results in classification and semantic segmentation tasks compared to state-of-the-art methods.
\end{abstract}

\section{Introduction}

Point clouds are the simplest shape representation for 3D geometric data and are the raw output of many 3D acquisition devices, e.g., LiDAR scanners. Feature learning from point clouds is crucial for many applications, such as 3D object detection \cite{Chen_2017_CVPR, Qi_2018_CVPR, Zhou_2018_CVPR}, 3D object classification and segmentation \cite{Qi_2017_CVPR, NIPS2018_7362, Yue_2019, Thomas_2019_ICCV}, 3D shape generation and correspondence \cite{Groueix_2018_ECCV, NIPS2019_8962}, and motion estimation of driving scenes \cite{Wang_2018_CVPR}. Inspired by the great success of convolutional neural networks (CNN) in the fields of natural language processing (1-D), image classification (2-D), and radiographic image analysis (3-D) where underlying data are Euclidean structured, deep neural networks on point clouds have recently driven significant interests. Directly applying CNN on point clouds is a challenge since they are non-Euclidean structured and the number and orientation of each point's neighbors vary from one to another. An effective definition of convolutional operation analogous to that on Euclidean structured data is important for feature learning on point clouds.

Recently, many deep neural networks have been developed to handle point clouds and achieved promising results. PointNet \cite{Qi_2017_CVPR} designed a deep learning framework that operates on each point individually and aggregates all individual point features to a global signature. However, the local structure of point clouds is not fully exploited. DGCNN \cite{Yue_2019} proposed a neural network module that explicitly constructs a local graph and learns embeddings for the edges to capture local geometric structure. However, DGCNN \cite{Yue_2019} applies isotropic filters over each point's neighbors, which limits the representation power compared to CNN.

Several works have introduced anisotropic filters for point cloud processing. PointCNN \cite{NIPS2018_7362} proposed to learn an $\mathcal{X}$-transformation from the input points to weight and permutate each point's neighbors into a latent and potentially canonical order. KPConv \cite{Thomas_2019_ICCV} presented a convolutional operation that weights each point's neighbors depending on the Euclidean distances to a set of predefined or deformable kernel points. Subsequently, these methods apply anisotropic filters on the resampled neighbors to extract features for point clouds. However, these methods do not explicitly explain why the transformation learned from the input points or the weighting matrix calculated from the kernel points can rearrange the irregular point clouds into regular and structured data so that anisotropic filters are able to take effect.

In this paper, we propose a \textbf{p}ermutable \textbf{a}n\textbf{i}sotropic convolutional operator (\emph{PAI-Conv}) for point clouds. Considering the neighbors of points can be in any directions for point clouds, we generate a set of kernel points that are evenly distributed on the surface of a sphere using Fibonacci lattice \cite{dixon1987mathographics} as a reference of the canonical order. For each point, we calculate the dot product between each point's neighbors and the kernel points and then apply sparsemax \cite{martins2016softmax} to obtain a sparse soft-permutation matrix. No weight is to learn for the sparse soft-permutation matrix. The idea of dot product with kernel points is similar to the dot product with keys in Transformer  \cite{NIPS2017_7181} that has demonstrated great success in natural language processing (NLP). Mathematically, using dot product attention and sparsemax resamples and reorders the local neighbors so that the convolutional neighbors follow the canonical order determined by the orientations of kernel points. Then similar to CNN, we apply shared anisotropic filters on the convolutional neighbors to extract local features on point clouds. PAI-Conv is designed to explicitly resample and reorder the neighbors and is physically meaningful.

PAI-Conv uses dot-product attention to rearrange the local neighbors, which mainly relies on each neighbor's orientation instead of the distance. Thus, PAI-Conv is robust to unevenly distributed point clouds and can work well with random point sampling method instead of using other complex sampling approaches \cite{NIPS2017_7095, NIPS2018_7362, GroWieLen18, Dovrat_2019_CVPR, abid2019concrete, Wu_2019_CVPR}. Thus, PAI-Conv is efficient and can be used in large-scale point clouds. We use PAI-Conv as a basic building module in existing frameworks and achieve state-of-the-art performance in classification and segmentation on several datasets. The main contributions of this paper are summarized below:

1) Considering the irregularity of point clouds, we propose a \textbf{p}ermutable \textbf{a}n\textbf{i}sotropic convolution (\emph{PAI-Conv}) for representation learning from point clouds. PAI-Conv uses dot-product attention and sparsemax to calculate a soft-permutation matrix for each point according to a set of fixed kernel points to rearrange the local neighbors in canonical order. The permutation matrix is similar to the attention matrix in Transformer \cite{NIPS2017_7181} that is widely used in NLP. To our best knowledge, this is the first work to learn permutation arrangement for point clouds, in contrast to the recent transformation matrix \cite{NIPS2018_7362} or weighting matrix \cite{Thomas_2019_ICCV} where no constraint is used to regulate the resampled neighbors.

2) PAI-Conv is robust and work well with the efficient random point sampling method. PAI-Conv layer is orthogonal to the basic convolutional layer for point clouds and can be readily integrated into existing pipelines \cite{Yue_2019, hu2019randla} for large-scale point clouds, by replacing the conv layer with PAI-Conv.

3) Extensive experiments show that our approach achieves state-of-the-art performance regarding with both accuracy and efficiency for classification and segmentation on point clouds.

\section{Related Work}

\textbf{Graph neural networks} Closely related to point clouds, graph data are also irregular but with edge connections. Graph neural networks are approaches to generalize CNN to irregular graph-structured data. Some approaches define graph convolutions based on a node's spatial relations.
GraphSage \cite{NIPS2017_6703} samples a fixed number of neighbors and aggregates neighboring features for each node. GAT \cite{velickovic2018graph} adopts attention mechanisms to learn the relative weights between two connected nodes. MoNet \cite{Monti_2017_CVPR} introduces node pseudo-coordinates to determine the relative position between a node and its neighbors and assigns different weights to the node's neighbors. FeaStNet \cite{Verma_2018_CVPR} proposed a graph-convolution operator that learns a weighting matrix dynamically computed from features.

\begin{figure*}[tb!]
    \centering
    \includegraphics[width=1.\textwidth]{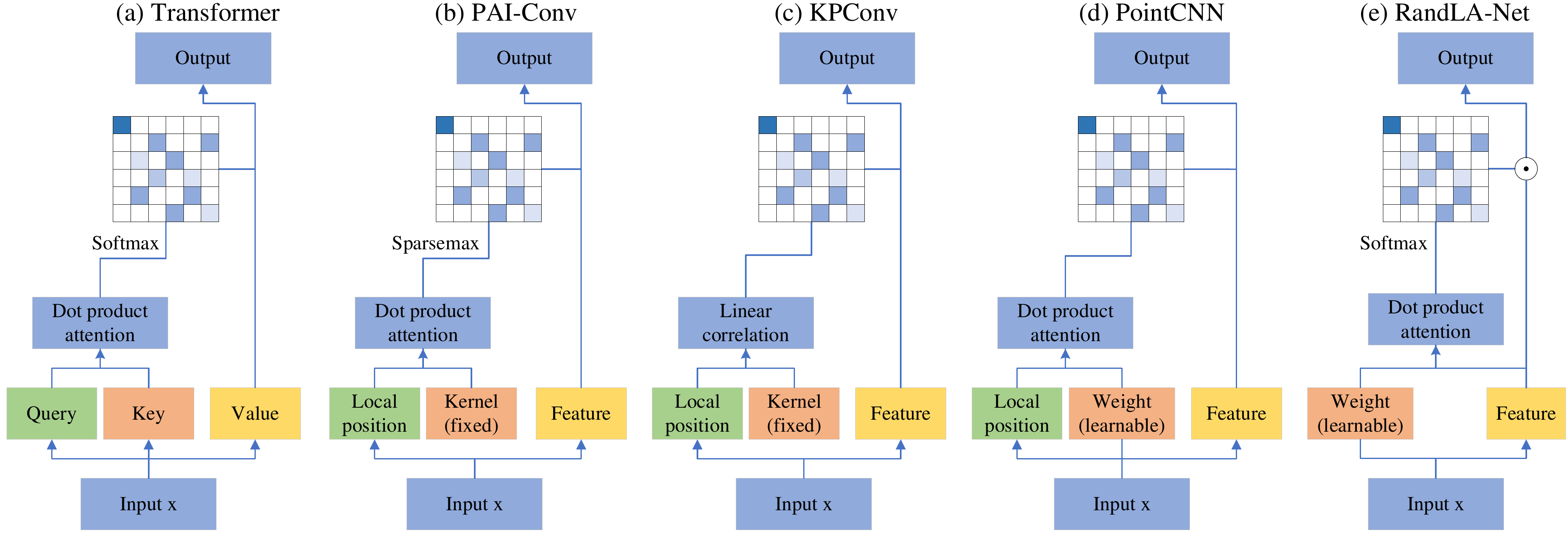}
\caption{Architecture comparison to existing methods.}
\label{fig:pai-gcn-compare}
\end{figure*}

\textbf{Transformer} Transformer models \cite{NIPS2017_7181} have demonstrated great success across a wide range of applications, especially for natural language processing (NLP) \cite{devlin2018bert}. In Transformer models for NLP, a sentence (i.e., a sequence of words) is considered as a fully-connected graph and Transformer can be considered as one kind of graph neural networks. As shown in Fig. \ref{fig:pai-gcn-compare}, the dot-product attention is both used in PAI-Conv and Transformer. Compared to Transformer, PAI-Conv uses local neighboring position as the query, a set of fixed kernel points as the key, point features as the value, and sparsemax as the activation function.

\textbf{Point neural networks} PointNet \cite{Qi_2017_CVPR} proposed an efficient architecture that directly applies multi-layer perceptrons (MLP) on each point. However, PointNet does not capture local structures from point clouds. Many neighboring feature pooling methods \cite{Li_2018_CVPR, zhao2019pointweb, Zhang_2019_ICCV} have been presented. PointNet++ \cite{NIPS2017_7095} proposed to sample points in a hierarchical fashion. DGCNN \cite{Yue_2019} proposed a convolutional operation that generates edge features to describe the relationships between a point and its neighbors. PointCNN \cite{NIPS2018_7362} presented a method to learn an $\mathcal{X}$-transformation as a function of input points. RandLA-Net \cite{hu2019randla} used an attention mechanism to learn local features instead of applying filters for convolutional operation. Besides, many kernel-based convolutional methods \cite{Su_2018_CVPR, Hua_2018_CVPR, Lei_2019_CVPR, Komarichev_2019_CVPR, Lan_2019_CVPR, Mao_2019_ICCV} have been devised. KPConv \cite{Thomas_2019_ICCV} showed a convolution that takes radius neighbors as input and processes them with weights spatially located by a set of kernel points.

Instead of calculating a soft-permutation matrix from the kernel points using the dot-product attention, KPConv~\cite{Thomas_2019_ICCV} computes a weighting matrix by linear correlation with an extra hyperparameter $\sigma$. The performance can be sensitive to $\sigma$ since the input density may not be constant and a small variation of $\sigma$ may significantly affect the weighting matrix. Furthermore, the dot product in PAI-Conv is more efficient than the linear correlation in KPConv. At last, different from weighting matrices, PAI-Conv applies sparsemax to obtain sparse soft-permutation matrices to explicitly permute each point's neighbors into canonical order. For PointCNN \cite{NIPS2018_7362}, instead of using kernel points to determine the canonical order, PointCNN learns transformation matrices from each point's neighbors using MLP. The weights of MLP can be considered as learnable kernel points, where the idea is similar to deformable KPConv \cite{Thomas_2019_ICCV}. However, it is difficult to stabilize the learnable kernel points when they are trained along with the whole network. Furthermore, the learnable matrix is not constrained by any form to act as a transformation matrix to permute the point's neighbors into canonical order.


\section{Our Approach}

Inspired by the conventional convolution on Euclidean structured data, we propose a permutable anisotropic convolutional operation (PAI-Conv) on point clouds. Instead of directly applying anisotropic filters on each point's neighbors, we rearrange each point's neighbors using a soft permutation matrix that is calculated based on each point's neighbors and a set of fixed kernel points. Then similar to CNN, we apply anisotropic filters to extract features from point clouds.

\subsection{Permutable anisotropic convolution}

Consider a point cloud $\bm{\mathcal{P}}=\{\vp_1, \ldots, \vp_N\}$ with per-point features $\vf \in \R^{D}$, where $D$ and $N$ represents the feature dimension and number of points, respectively. Each point contains 3D coordinates $\vp_i=[x_i, y_i, z_i]^\top$ in the Euclidean space. Point features can include RGB color or intermediate learned features in a deep neural network. For each point, its neighboring points are gathered by the simple K-nearest neighbors (KNN) algorithm based on the point-wise Euclidean distances for efficiency. We denote the $i$th point's $K$-nearest neighbors as $\bm{\mathcal{N}}_i = \{\vp_i, \vp_{i, 1}, \ldots, \vp_{i, K-1}\}$.

\begin{figure*}[t]
    \centering
    \includegraphics[width=0.9\textwidth]{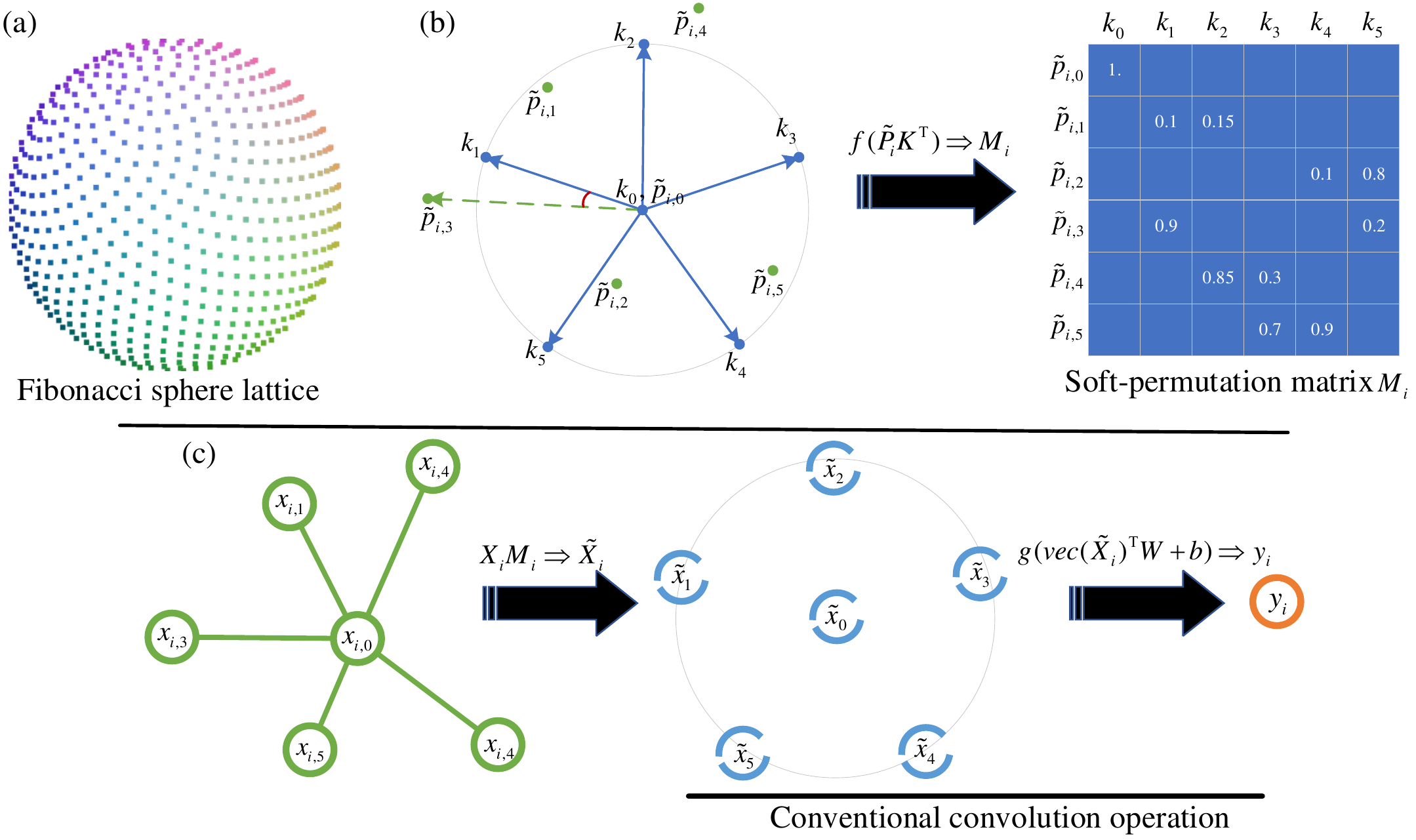}
\caption{Permutable anisotropic convolutional operation (PAI-Conv). (a) Visualization of Fibonacci sphere lattice for kernel points, $\{\vk_0, \vk_1, \ldots, \vk_{L-1}\}$. (b) Generating soft-permutation matrix, $\mM_i$, explained in 2-D. $\{\tilde{\vp}_{i,0}, \tilde{\vp}_{i,1}, \ldots, \tilde{\vp}_{i,K-1}\}$ are the local neighboring positions of point $\vp_i$. The permutation matrix $\mM_i$ is calculated by the dot product of the local neighboring positions and the kernel points followed by sparsemax $f(\cdot)$. (c) $\{\vx_{i,0}, \vx_{i,1}, \ldots, \vx_{i, 5}\}$ are the point features of point $\vp_i$'s neighbors. PAI-Conv uses the soft-permutation matrix, $\mM_i\in\R^{K \times L}$, to resample the node's neighbors. Then, a conventional convolution operation with anisotropic filters followed by ELU $g(\cdot)$ is performed. The anisotropic filters have weights $\mW\in\R^{(D_{in}\cdot L)\times D_{out}}$ and bias $\vb\in\R^{D_{out}}$.}
\label{fig:pai-gcn}
\end{figure*}

\textbf{Soft-permutaton matrix} We generate a set of kernel points by mapping the Fibonacci lattice \cite{dixon1987mathographics} onto the surface of a sphere via equal-area projection so that the kernel points are evenly distributed on a sphere, denoted as $\bm{\mathcal{K}} = \{\vk_0, \vk_1, \ldots, \vk_{L-1}\}$, where $\vk_0=[0, 0, 0]^\top$ is at the origin and $L$ is the number of kernel points, as shown in Figure \ref{fig:pai-gcn}a. For the $i$th point, we obtain the local neighboring positions of the neighbors as $\tilde{\vp}_{i,j}=\vp_i - \vp_{i,j}$, where $\vp_{i,j}\in\bm{\mathcal{N}}_i$ and $\tilde{\vp}_{i,0}=[0, 0, 0]^\top$ is at the origin. As shown in Figure \ref{fig:pai-gcn}b, the soft-permutation matrix is simply calculated by the dot product between the local neighboring positions and the kernel points followed by sparsemax \cite{martins2016softmax}, expressed as
\begin{align}\label{eq:permutation}
\mM_i = f(\tilde{\mP}_i\mK^\top),
\end{align}
where $\tilde{\mP}_i\in\R^{K\times 3}, \mK\in\R^{L\times 3}, \mM_i\in\R^{K\times L}$, and $f(\cdot)$ is sparsemax. For each kernel point, the dot product ensures a local neighboring position with a smaller angle to the kernel point has a larger weight. Sparsemax ensures the soft-permutation matrix is sparse and only those points with small angles to a kernel point are selected otherwise the weight is zero. For example, as shown in Figure \ref{fig:pai-gcn}b, $\tilde{\vp}_{i, 3}$ and $\tilde{\vp}_{i, 1}$ have small angles to $\vk_1$ and are assigned with weight of 0.9 and 0.1, respectively. Other points with larger relative angles are assigned with weight of zero. Note that, since $\vk_0=\tilde{\vp}_{i,0}=[0, 0, 0]^\top$, the dot product of these two vectors is zero. We need to set $\mM[0,0]=1$ to make the point itself be selected as the first point in the resample convolutional neighbors.

\textbf{Point feature} Inspired by RandLA-Net \cite{hu2019randla}, we first encode the relative point position as
\begin{align}\label{eq:relative}
\vr_{i,j} = MLP(\vp_i \oplus (\vp_i - \vp_{i,j}) \oplus \|\vp_i - \vp_{i,j}\|),
\end{align}
where $\oplus$ is the concatenation operation and $\|\cdot\|$ calculates the Euclidean distance between the neighbor s and the center point. Then, the point feature is obtained by concatenating the relative point position and the per-point features as follows,
\begin{align}\label{eq:feature}
\vx_{i,j} = \vr_{i,j} \oplus \vf_{i,j},
\end{align}
where $\vf_{i,j}$ is the RGB color or intermediate learned features as mentioned above, $\vx_{i,j}\in\R^{D_{in}}$, and $D_{in}$ is the feature dimension. For each point, we construct $\mX_i = \{\vx_{i,0} ,\vx_{i,1}, \ldots, \vx_{i,K-1}\}\in \R^{D_{in}\times K}$ for the convolutoinal operation.

Since the order and orientation of each point's neighbors vary from one to another, directly applying an anisotropic filter on unordered neighbors diminishes the representation power. While training, the anisotropic filter might struggle to adapt to the large variability of the unordered coordinate systems and the possibility of learning rotation invariant filter increases. In this paper, we resample each point's neighbors using the soft-permutation matrix in Eq. (\ref{eq:permutation}). The resampled convolutional neighbors of each point can be obtained by
\begin{align}\label{eq:resample}
\tilde{\mX}_i = \mX_i \mM_i,
\end{align}
where $\tilde{\mX}_i\in \R^{D_{in} \times L}$. Since the point's neighbors are rearranged according to the canonical order of the fixed kernel points, we can apply a shared anisotropic filter on each point of a point cloud. This operation is the same as the conventional convolution and can be expressed as
\begin{align}\label{eq:conv}
\vy_i = g(\text{vec}(\tilde{\mX}_i)^\top \mW + \vb),
\end{align}
where $\mW\in\R^{(D_{in}\cdot L) \times D_{out}}$ includes $D_{out}$ anisotropic filters, $\vb\in\R^{D_{out}}$ is the bias, $\vy_i\in\R^{D_{out}}$ is the output feature point corresponding to the input feature point $\vx_i\in\R^{D_{in}}$, $vec(\cdot)$ is a vectorization function which converts a matrix into a column vector, and $g(\cdot)$ is an activation function, e.g., ELU \cite{clevert2015fast}, to introduce non-linearity.

\subsection{Properties}

Intuitively, \textbf{anisotropic filter} and \textbf{permutation invariance} are two mutually exclusive properties. Using anisotropic filter requires each point's neighbors are sorted in a certain way while permutation invariance means we can randomly change the order of each point's neighbors. The proposed PAI-Conv resolves these two incompatible properties by introducing soft-permutation matrices so that the initial order of each point's neighbors is not important. PAI-Conv achieves permutation invariance both for the order of points in a point cloud and the order of each point's neighbors. KNN makes sure each point has the same neighbors even we reshuffle the order of points in a point cloud. The dot-product attention and sparsemax make sure the calculated soft-permutation matrix can rearrange the neighbors in the canonical order defined by the kernel points even we reshuffle the order of each point's neighbors. Since each point's neighbors follow the same canonical order, we can apply anisotropic filters to extract local features with high representation power for point clouds.

\section{Experiments}
We evaluate PAI-Conv for two different tasks: classification and semantic segmentation. Ablation tests are conducted to demonstrate the effectiveness of the architecture design.

\subsection{Classification}

We evaluate our model on the ModelNet40 \cite{Wu_2015_CVPR} for the classification task. The dataset contains 12,311 meshed CAD models from 40 clauses: 9,843 models for training and 2,468 models for testing. The Mean Accuracy (MA) and Overall Accuracy (OA) of all classes are used as the standard metrics. We test PAI-Conv in two different data sampling methods: uniformly sampling and randomly sampling.

\subsubsection{Uniformly sampling data}\label{sec:uniform}

\emph{Data.} Following the experimental setting of \cite{Qi_2017_CVPR, Yue_2019}, we uniformly sample 2,048 points on mesh faces according to the face area and normalize them into a unit sphere. The training data are augmented on-the-fly by randomly scaling points in the range of [2/3, 3/2], translating points in the range of [-0.2, 0.2], and jittering points by Gaussian noise with zero mean and 0.01 standard deviation.

\emph{Architecture.} We adopt the network architecture from \cite{Yue_2019} and replace the EdgeConv with our proposed PAI-Conv, where the size of kernel points is $L=32$ and each point's neighbors is $K=40$. First, soft-permeation matrices for each point are computed. We use four PAI-Conv layers with channels of [64, 64, 128, 256] to extract geometric features. Since no down-sampling is applied in the network, the four PAI-Conv layers share the same neighboring indexes and soft-permutation matrices. Shortcut connections are included to extract multi-scale features and one shared MLP layer [2048] is used to aggregate multi-scale features. Then, we use max/sum pooling along the dimension of point cloud's size to get the global feature. At last, two fully-connected layers with channels of [512, 40] and two dropout layers with 0.5 probability are used to finally predict a 3D shape's class.

\emph{Training.} We use SGD with initial learning rate 0.1 and schedule it to be ended at 0.01 using cosine annealing method \cite{Loshchilov2016} for 250 epochs. The batch size is 16 and the momentum is 0.9.

\begin{table}
  \caption{Classification on ModelNet40. MA: mean accuracy; OA: overall accuracy.}
  \label{tb:cls}
  \centering
  \footnotesize
  \begin{tabular}{lcclcc}
 \toprule
    & MA & \multicolumn{1}{c|}{OA} & &  MA & OA \\
    \cmidrule(lr){1-3} \cmidrule(lr){4-6}
    3DShapeNets \cite{Wu_2015_CVPR} & 77.3& 84.7 & PointNet \cite{Qi_2017_CVPR} & 86.0 & 89.2 \\
    VoxNet \cite{Maturana2015} & 83.0 & 85.9 & PointNet++ \cite{NIPS2017_7095} & - & 90.7 \\
    Subvolume \cite{Qi_2016_CVPR} & 86.0 & 89.2 & Kd-net \cite{Klokov_2017_ICCV} & - & 90.6 \\
    VRN \cite {Brock2016} & - & 91.3 & PointCNN \cite{NIPS2018_7362} & 88.1 & 92.2 \\
    ECC \cite{Simonovsky_2017_CVPR} & 83.2 & 87.4 & PCNN \cite{Atzmon2018} & - & 92.3 \\
    SpecGCN \cite{Wang_2018_ECCV} & - & 91.5 & KPConv \cite{Thomas_2019_ICCV} & - & 92.9  \\
    \cmidrule(lr){1-3} \cmidrule(lr){4-6}
    \multicolumn{3}{c|}{\bf Uniformly Sampling} & \multicolumn{3}{c}{\bf Randomly Sampling}  \\
    DGCNN \cite{Yue_2019} & \textbf{90.7} & \multicolumn{1}{c|}{\textbf{93.5}} & DGCNN \cite{Yue_2019} & 89.0 & 92.4 \\
    RandLA-Net \cite{hu2019randla} & 84.7 & \multicolumn{1}{c|}{89.8} & RandLA-Net \cite{hu2019randla} & 86.1 & 90.7 \\
  \textbf{PAI-Conv (Ours)} & 90.0 & \multicolumn{1}{c|}{93.2} & \textbf{PAI-Conv (Ours)} & \textbf{90.5} & \textbf{93.2} \\
  \bottomrule
  \end{tabular}
\end{table}

\emph{Results.} Table \ref{tb:cls} shows the results on ModelNet40. PAI-Conv is competitive against state-of-the-art methods. When the point clouds are uniformly sampled, PAI-Conv achieves the results that are only slightly lower than DGCNN \cite{Yue_2019}. Since the point clouds are uniformly sampled, the neighbors of each point are evenly distributed, making anisotropic filters in PAI-Conv redundant. Thus, the advantage of PAI-Conv over DGCNN is not well exploited. However, uniformly sampling data is computationally expensive, which is not practical for large scale point clouds in applications.

\subsubsection{Randomly sampling data}

Instead of relying on expensive sampling techniques or computationally heavy pre/post- processing steps, we randomly sample 8192 points from each point cloud. The same network architecture and training procedure as in Section \ref{sec:uniform} are used. Through the four PAI-Conv layers, the point cloud is down-sampled randomly with the ratios of [4, 4, 2, 2]. Since the point cloud is down-sampled, the neighboring indexes and soft-permutation matrices are calculated for each PAI-Conv layer.

Table \ref{tb:cls} also shows the classification results on ModelNet40 when the point clouds are randomly sampled. PAI-Conv achieves better results than DGCNN \cite{Yue_2019}. Randomly sampling data is efficient and can easily be applied to large-scale point clouds. The total time and memory consumption comparisons of different sampling methods are demonstrated in Figure 5 of \cite{hu2019randla}. We re-implement RandLA-Net \cite{hu2019randla} in both uniformly and randomly sampling methods. PAI-Conv outperforms RandLA-Net in a large margin for the classification task.

\subsection{Semantic segmentation}

In this section, we perform semantic segmentation on three large-scale public datasets as follows:

\textbf{SemanticKITTI dataset} \cite{Behley_2019_ICCV} The SemanticKITTI consists of 22 sequences of densely annotated LiDAR scans: sequence 00-07 and 09-10 with 19,130 scans for training, sequence 08 with 4,071 scans for validation, and sequence 11-21 with 20,351 scans for testing. Each scan is a large-scale point cloud with around $10^5$ points and up to $160\times160\times20$ meters in 3D space. The SemanticKITTI dataset only includes 3D coordinates without other feature information. The Mean Intersection-over-Union (mIoU) score over 19 categories is used as the standard metric.

\textbf{Semantic3D dataset} \cite{hackel2017isprs} The Semantic3D consists of 30 terrestrial laser scans: 15 for training and 15 for testing. Each point cloud has up to $10^8$ points and takes up to $160\times240\times30$ meters in 3D space. The Semantic3D dataset includes 3D coordinates and RGB information that are used as the input for training and testing. The mIoU and OA of all 8 classes are used as the standard metrics.

\textbf{S3DIS dataset} \cite{Armeni2018} The S3DIS consists of 6 large-scale indoor areas from three different buildings: 5 for training and 1 for testing using the standard 6-fold cross-validation. Each point cloud is a medium-sized single room around $20\times15\times 5$ meters in 3D space. The S3DIS dataset includes 3D coordinates and RGB information that are used as the input for training and testing. The mIoU, mean class Accuracy (mAcc), and OA of the total 13 classes are used as the standard metrics.

\emph{Architecture.} We adopt the network architecture from \cite{hu2019randla} and replace the LocSE module with our proposed PAI-Conv, where the size of kernel points is $L = 16$ and each point’s neighbors is $K = 16$. The network is an encoder-decoder architecture with skip connections. First, the input point cloud is fed to an MLP layer to extract per-point feature with output channel 8. Then, five encoder and five decoder networks are connected by an MLP layer. Each encoder layer consists of a residual block with a PAI-Conv layer and a random sampling operation. Through the five encoder layers, the point cloud is down-sampled with ratios of [4, 4, 4, 4, 2] and the feature dimension is increased with channels of [16, 64, 128, 256, 512]. In each decoder layer, the point feature is up-sampled by a nearest-neighbor interpolation and concatenated with the intermediate feature from the corresponding encoder layer through a skip connection. The concatenated feature is fed to an MLP layer. At last, three fully-connected layers with channels of [64, 32, $num\_class$] and a dropout with 0.5 probability are used to finally predict the semantic label of each point.

\emph{Training.} We use Adam optimizer with an initial learning rate of 0.01 and schedule the learning rate with decay 0.95 for 100 epochs. The training batch size is 6.

\begin{table}[tb!]
  \caption{Quantitative results on SemanticKITTI \cite{Behley_2019_ICCV} (best performance in bold).}
  \label{tb:segkitti}
  \centering
  \scriptsize
  \resizebox{1.0\textwidth}{!}{
  \begin{tabular}{p{0.2cm}p{2.1cm}P{0.18cm}P{0.18cm}P{0.18cm}P{0.18cm}P{0.18cm}P{0.18cm}P{0.18cm}P{0.18cm}P{0.18cm}P{0.18cm}P{0.18cm}P{0.18cm}P{0.18cm}P{0.18cm}P{0.18cm}P{0.18cm}P{0.18cm}P{0.18cm}P{0.18cm}P{0.18cm}P{0.18cm}}
    \toprule
     Size&Methods&\rotatebox[origin=l]{90}{\textbf{mIoU(\%)}} & \rotatebox[origin=l]{90}{Params(M)} & \rotatebox[origin=l]{90}{road} & \rotatebox[origin=l]{90}{sidewalk} & \rotatebox[origin=l]{90}{parking} & \rotatebox[origin=l]{90}{other-ground} & \rotatebox[origin=l]{90}{building} & \rotatebox[origin=l]{90}{car} & \rotatebox[origin=l]{90}{truck} & \rotatebox[origin=l]{90}{bicycle} & \rotatebox[origin=l]{90}{motorcycle} & \rotatebox[origin=l]{90}{other-vehicle} & \rotatebox[origin=l]{90}{vegetation} & \rotatebox[origin=l]{90}{trunk} & \rotatebox[origin=l]{90}{terrain} & \rotatebox[origin=l]{90}{person} & \rotatebox[origin=l]{90}{bicyclist} & \rotatebox[origin=l]{90}{motorcyclist} & \rotatebox[origin=l]{90}{fence} & \rotatebox[origin=l]{90}{pole} & \rotatebox[origin=l]{90}{traffic-sign} \\
    \hline
    \multirow{5}{*}{\rotatebox[origin=c]{90}{\parbox{0.8cm}{64$\times$2048\\pixels}}}&SqueezeSeg \cite{Wu2018} &29.5 &1 &85.4 &54.3 &26.9 &4.5 &57.4 &68.8 &3.3 &16.0 &4.1 & 3.6 &60.0 &24.3 &53.7 &12.9 &13.1 &0.9 &29.0 &17.5 & 24.5 \\
    &SqueezeSegV2 \cite{Wu2019} &39.7 &1 &88.6 &67.6 &45.8 &17.7 &73.7 &81.8 &13.4 &18.5 &17.9 &14.0 &71.8 &35.8 &60.2 &20.1 &25.1 &3.9 &41.1 &20.2 &36.3 \\
    &DarkNet21Seg \cite{Behley_2019_ICCV} &47.4 &25 &91.4 &74.0 &57.0 &26.4 &81.9 &85.4 &18.6 &26.2 &26.5 &15.6 &77.6 &48.4 &63.6 &31.8 &33.6 &4.0 &52.3 &36.0 &50.0 \\
    &DarkNet53Seg \cite{Behley_2019_ICCV} &49.9 &50 &\textbf{91.8} &74.6 &64.8 &\textbf{27.9} &84.1 &86.4 &25.5 &24.5 &32.7 &22.6 &78.3 &50.1 &64.0 &36.2 &33.6 &4.7 &55.0 &38.9 &52.2 \\
    &RangeNet53++ \cite{Milioto2019} &52.2 &50 &\textbf{91.8} &\textbf{75.2} &\textbf{65.0} &27.8 &\textbf{87.4} &91.4 &25.7 & 25.7 &\textbf{34.4} &23.0 &80.5 &55.1 &64.6 &38.3 & 38.8 &4.8 &58.6 &47.9 &\textbf{55.9} \\
    \hline
    \multirow{7}{*}{\rotatebox[origin=c]{90}{50K pts}}&PointNet \cite{Qi_2017_CVPR}&14.6&3&61.6&35.7&15.8&1.4 & 41.4 & 46.3 & 0.1 & 1.3 & 0.3 & 0.8 & 31.0 & 4.6 & 17.6 & 0.2 & 0.2 & 0.0 & 12.9 & 2.4 & 3.7 \\
    &SPG \cite{Landrieu_2018_CVPR} &17.4 & 0.25&45.0 &28.5 &0.6 &0.6 &64.3 &49.3 &0.1 &0.2 &0.2 &0.8 &48.9 &27.2 &24.6 &0.3 &2.7 &0.1 &20.8 &15.9 &0.8 \\
    &SPLATNet \cite{Su_2018_CVPR} &18.4 &0.8 &64.6 &39.1 &0.4 &0.0 &58.3 &58.2 &0.0 &0.0 &0.0 &0.0 &71.1 &9.9 &19.3 &0.0 &0.0 &0.0 &23.1 &5.6 &0.0\\
    &PointNet++ \cite{NIPS2017_7095} &20.1 &6 &72.0 &41.8 &18.7 &5.6 &62.3 &53.7 &0.9 &1.9 &0.2 &0.2 &46.5 &13.8 &30.0 &0.9 &1.0 &0.0 &16.9 &6.0 &8.9 \\
    &TangentConv \cite{Tatarchenko_2018_CVPR} &40.9 &0.4 &83.9 &63.9 &33.4 &15.4 &83.4 &90.8 &15.2 &2.7 &16.5 &12.1 &79.5 &49.3 &58.1 &23.0 &28.4 &\textbf{8.1} &49.0 &35.8 &28.5 \\
    &RandLA-NetV1 \cite{hu2019randla} &50.3 & 0.95 &90.4 &67.9 &56.9 &15.5 &81.1 &94.0 &42.7 &19.8 &21.4 &38.7 &78.3 &60.3 &59.0 &47.5 &\textbf{48.8} &4.6 &49.7 &44.2 &38.1 \\
    &RandLA-NetV3 \cite{hu2019randla} &\textbf{53.9} &1.24 &90.7 &73.7 &60.3 &20.4 &86.9 &\textbf{94.2} &40.1 &\textbf{26.0} &25.8 &\textbf{38.9} &81.4 &61.3 &\textbf{66.8 } &49.2 &\textbf{48.2} &7.2 &56.3 &\textbf{49.2} &47.7 \\
    &\textbf{PAI-Conv (ours)} &\textbf{53.9} & 9.26 & 90.1 & 73.5 &58.6 &26.8 &87.8 &94.0 &\textbf{44.5} &25.5 &30.0 &36.2 &\textbf{81.9} &\textbf{61.5} &65.4 & 46.1 &45.9 &3.5 &\textbf{57.7} &49.0 &45.8\\
    \bottomrule
  \end{tabular}
  }
  \vspace{-1em}
\end{table}

\begin{table}[tb!]
  \caption{Quantitative results on Semantic3D \cite{hackel2017isprs} (best performance in bold).}
  \label{tb:segSemantic3D}
  \centering
  \footnotesize
  \begin{tabular}{p{2.7cm}P{0.9cm}P{0.6cm}P{0.6cm}P{0.7cm}P{0.5cm}P{0.5cm}P{0.85cm}P{0.6cm}P{0.8cm}P{0.5cm}}
    \toprule
    &\footnotesize mIoU(\%)&\footnotesize OA(\%) &\footnotesize man-made. &\footnotesize natural. &\footnotesize high veg. &\footnotesize low veg. &\footnotesize buildings &\footnotesize hard scape &\footnotesize scanning art. &\footnotesize car \\
    \hline
    SnapNet \cite{Boulch2017} & 59.1 & 88.6 & 82.0 & 77.3 & 79.7 & 22.9 & 91.1 & 18.4 & 37.3 & 64.4 \\
    SEGCloud \cite{Tchapmi2017} & 61.3 & 88.1 & 83.9 & 66.0 & 86.0 & 40.5 & 91.1 & 30.9 & 27.5 & 64.3 \\
    RF\_MSSF \cite{Thomas2018} & 62.7 & 90.3 & 87.6 & 80.3 & 81.8 & 36.4 & 92.2 & 24.1 & 42.6 & 56.6 \\
    MSDeepVoxNet \cite{Roynard2018} & 65.3 & 88.4 & 83.0 & 67.2 & 83.8 & 36.7 & 92.4 & 31.3 & 50.0 & 78.2 \\
    ShellNet \cite{Zhang_2019_ICCV} & 69.3 & 93.2 & 96.3 & 90.4 & 83.9 & 41.0 & 94.2 & 34.7 & 43.9 & 70.2 \\
    GACNet \cite{Wang_2019_CVPR} & 70.8 & 91.9 & 86.4 & 77.7 & \textbf{88.5} & \textbf{60.6} & 94.2 & 37.3 & 43.5 & 77.8 \\
    SPG \cite{Landrieu_2018_CVPR} & 73.2 & 94.0 & \textbf{97.4} & \textbf{92.6} & 87.9 & 44.0 & 83.2 & 31.0 & 63.5 & 76.2 \\
    KPConv \cite{Thomas_2019_ICCV} & 74.6 & 92.9 & 90.9 & 82.2 & 84.2 & 47.9 & 94.9 & 40.0 & \textbf{77.3} & \textbf{79.7} \\
    RandLA-NetV1 \cite{hu2019randla} & 76.0 & 94.4 & 96.5 & 92.0 & 85.1 & 50.3 & 95.0 & 41.1 & 68.2 & 79.4 \\
    RandLA-NetV3 \cite{hu2019randla} & \textbf{77.4} & \textbf{94.8} & 95.6 & 91.4 & 86.6 & 51.5 & \textbf{95.7} & \textbf{51.5} & 69.8 & 76.8 \\
    \textbf{PAI-Conv (Ours)} & 76.4& 94.4& 96.4& 90.4& 86.7& 49.6& 95.6& 45.4&69.0&78.4 \\
    \bottomrule
  \end{tabular}
  \vspace{-1em}
\end{table}

\begin{table}[tb!]
  \caption{\footnotesize Quantitative results on the S3DIS dataset \cite{Armeni2018} (6-fold cross validation, best performance in bold).}
  \label{tb:seg3DIS}
  \centering
  \footnotesize
  \begin{tabular}{p{1.2cm}P{0.4cm}P{0.4cm}P{0.4cm}P{0.4cm}P{0.4cm}P{0.4cm}P{0.4cm}P{0.4cm}P{0.4cm}P{0.4cm}P{0.4cm}P{0.4cm}P{0.4cm}P{0.4cm}P{0.4cm}}
    \toprule
    &\scriptsize\rotatebox[origin=l]{90}{\parbox{1.4cm}{PointNet\\ \cite{Qi_2017_CVPR}}} &\scriptsize\rotatebox[origin=l]{90}{\parbox{1.4cm}{PointNet++\\ \cite{NIPS2017_7095}}} &\scriptsize\rotatebox[origin=l]{90}{\parbox{1.4cm}{DGCNN\\ \cite{Yue_2019}}} &\scriptsize\rotatebox[origin=l]{90}{\parbox{1.4cm}{3P-RNN \\ \cite{Ye_2018_ECCV}}} &\scriptsize\rotatebox[origin=l]{90}{\parbox{1.4cm}{RSNet \\\cite{Huang_2018_CVPR}}} &\scriptsize\rotatebox[origin=l]{90}{\parbox{1.4cm}{SPG \\ \cite{Landrieu_2018_CVPR}}} &\scriptsize\rotatebox[origin=l]{90}{\parbox{1.4cm}{LSANet \\\cite{chen2019lsanet}}} &\scriptsize\rotatebox[origin=l]{90}{\parbox{1.4cm}{PointCNN \\\cite{NIPS2018_7362}}} &\scriptsize\rotatebox[origin=l]{90}{\parbox{1.4cm}{PointWeb\\ \cite{Zhao_2019_CVPR}}} &\scriptsize\rotatebox[origin=l]{90}{\parbox{1.4cm}{ShellNet \\\cite{Zhang_2019_ICCV}}} &\scriptsize\rotatebox[origin=l]{90}{\parbox{1.4cm}{HEPIN \\\cite{Jiang_2019_ICCV}}}&\scriptsize\rotatebox[origin=l]{90}{\parbox{1.4cm}{KPConv \\ \cite{Thomas_2019_ICCV}}} &\scriptsize\rotatebox[origin=l]{90}{\parbox{1.4cm}{RandLA-Net \\V1 \cite{hu2019randla}}}
    &\scriptsize\rotatebox[origin=l]{90}{\parbox{1.4cm}{RandLA-Net \\V3 \cite{hu2019randla}}} &\scriptsize\rotatebox[origin=l]{90}{\parbox{1.4cm}{\textbf{PAI-Conv}\\ \textbf{(Ours)}}}\\
    \hline
    OA(\%) &78.6 &81.0 &84.1 &86.9 &- &85.5 &86.8 &88.1 &87.3 &87.1 &\textbf{88.2} &- & 87.2&88.0 & 87.8 \\
    mAcc(\%) &66.2 &67.1 &- &- &66.5 &73.0 &- &75.6 &76.2 &- &-& 79.1 & 81.5&\textbf{82.0} & 80.9\\
    mIoU(\%) &47.6 &54.5 &56.1 & 56.3 &56.5 &62.1 & 62.2 &65.4 &66.7 &66.8 &67.8 &\textbf{70.6} & 68.5&70.0 & 69.1\\
    \bottomrule
  \end{tabular}
\end{table}

\emph{Results.} Table \ref{tb:segkitti}, Table \ref{tb:segSemantic3D}, and Table \ref{tb:seg3DIS} present the quantitative results of different approaches on the SemanticKITTI, Semantic3D, and S3DIS dataset, respectively. PAI-Conv achieves on-par or better performance than state-of-the-art methods. Compared with the most recent RandLA-Net \cite{hu2019randla}, PAI-Conv outperforms the version one (V1) of RandLA-Net in all the three datasets and is slightly lower than the version three (V3) of RnadLA-Net in Semantic3D and S3DIS dataset. Note that, the small difference between RandLA-Net and PAI-Conv may be eliminated by repeatedly testing on the datasets. For SemanticKITTI dataset, two types of methods: point-based (\emph{50K pts}) and projection-based (\emph{64*2048 pixels}) are presented. DarkNet53Seg \cite{Behley_2019_ICCV} and RangeNet53++ \cite{Milioto2019} achieve better results for some objects. However, they require costly steps for the pre/post projection.

\begin{figure}[tb!]
\begin{minipage}{1\textwidth}
\begin{minipage}[b]{0.51\textwidth}
\centering
\footnotesize
\captionof{table}{The computation time, network parameters, and maximum number of input points of different approaches for semantic segmentation on Sequence 08 of the SemanticKITTI \cite{Behley_2019_ICCV} dataset.}
\begin{tabular}{p{2.3cm}P{1cm}P{1.2cm}P{1.2cm}}
    \toprule
    & \makecell{time \\ (\emph{s})} & \makecell{para \# \\ (\emph{M})} & \makecell{point \# \\(\emph{M})}\\
    \hline
    PointNet \cite{Qi_2017_CVPR} &192 & 0.8 & 0.49 \\
    PointNet++ \cite{NIPS2017_7095} &9831 & 0.97 &0.98 \\
    PointCNN \cite{NIPS2018_7362} & 8142 & 11 & 0.05\\
    SPG \cite{Landrieu_2018_CVPR} & 43584 & \textbf{0.25} & -\\
    KPConv \cite{Thomas_2019_ICCV} & 717 & 14.9 & 0.54\\
    RandLA-Net \cite{hu2019randla} & 185 & 1.24 & 1.03\\
    \textbf{PAI-Conv (Ours)} & \textbf{176} & 9.26 & \textbf{1.51} \\
    \bottomrule
\end{tabular}
\label{tab:efficient}
\end{minipage}
\hfill\vline\hfill
\begin{minipage}[b]{0.42\textwidth}
\centering
\captionof{table}{Ablation tests of classification  on ModelNet40 for PAI-Conv. We uniformly sample 1,024 points for each 3D shape. MA: mean accuracy; OA: overall accuracy.}
\begin{tabular}{p{2.6cm}P{1cm}P{1cm}}
    \toprule
    & MA(\%) & OA(\%)\\[0.5ex]
    \hline
    w/o permutation & 88.0 &91.7 \\
    w/o sparsemax & 88.7 &92.3 \\
    w/ softmax & 86.4 &91.0\\
    w/ isotropic filter & 88.7 &91.1\\
    w/ random kernel & 88.9&92.5 \\
    w/ learnable kernel & 89.1 & 92.5\\
    \textbf{full model} & \textbf{90.0} & \textbf{92.9}\\
    \bottomrule
\end{tabular}
\label{tab:abltion}
\end{minipage}
\end{minipage}
\vspace{-1.5em}
\end{figure}

\subsection{Efficiency of PAI-Conv and ablation study}

In this section, we conduct experiments to demonstrate the efficiency of PAI-Conv. Following the same setting of \cite{hu2019randla}, we obtain the computation time, number of network parameters, and maximum number of input points of the proposed PAI-Conv for semantic segmentation on Sequence 08 of the SemanticKITTI \cite{Behley_2019_ICCV} dataset on the machine with an AMD 3700X @3.6GHz CPU and an NVIDIA RTX2080Ti GPU. Maximum number of input points means the number of 3D points each network can take as input in a single pass to infer per-point semantics. As shown in Table \ref{tab:efficient}, PAI-Conv achieves the shortest computation time and the largest number of input points. This is because, the same as RandLA-Net \cite{hu2019randla}, we use random point sampling method instead of other expensive sampling methods. PAI-Conv is more efficient than RandLA-Net because we use dot product on fixed kernel points and only use one PAI-Conv layer instead of two for each residual block.

We conduct ablation tests on the ModelNet40 \cite{Wu_2015_CVPR} to evaluate the effectiveness of the architecture design of PAI-Conv. We uniformly sample 1,024 points for each 3D shape. As shown in Table \ref{tab:abltion}, When without permutation matrices (\emph{w/o permutation}), the anisotropic filters are redundant since point clouds are irregular. When without sparsemax (\emph{w/o sparsemax}) or replacing sparsemax with softmax (\emph{w/ softmax}), the permutation matrices are not sparse enough and become weighting matrices that tend to average each point's neighbors. When replacing anisotropic filters with isotropic filters (\emph{w/ isotropic filter}), all the points are treated equally, which is the same as PointNet \cite{Qi_2017_CVPR}. When using random kernel points (\emph{w/ random kernel}) or learnable kernel points (\emph{w/ learnable kernel}, similar to PointCNN \cite{NIPS2018_7362}), the local coordinates that define the canonical order cannot well cover each point's local neighbors compared to the fixed kernel points generated using the Fibonacci lattice, since the neighbors of points in point clouds can be in any directions. Thanks to the design of PAI-Conv, the full model achieves the best results.

\section{Conclusion}

This paper proposes a permutable anisotropic convolutional operation (\emph{PAI-Conv}) for point clouds. PAI-Conv is a physically meaningful generalization of CNN from grid-structured data to irregular data. Dot-product attention and sparsemax are used to obtain sparse soft-permeation matrices so that each point's neighbors are rearranged in the canonical order of a set of fixed kernel points. Then anisotropic filters are applied to extract local features of point clouds. PAI-Conv can effectively cooperate with random point sampling method and can be applied in large-scale point clouds. Quantitative experiments demonstrate that PAI-Conv produces competitive results with state-of-the-art methods on classification and semantic segmentation for point clouds.

\section*{Broader Impact}
Point cloud data is becoming more and more ubiquitous.   Effective representation and learning can be of help in many industries like entertainment and sports. We shall also be careful such digitization to the physical world may also incur privacy issues for individuals as any of us may be recorded in the 3D point cloud form. Hence privacy-protection shall be always taken into under consideration when developing related technologies to this paper.

\bibliography{example_paper}

\begin{thebibliography}{10}\itemsep=-1pt

\bibitem{abid2019concrete}
Abubakar Abid, Muhammed~Fatih Balin, and James Zou.
\newblock Concrete autoencoders for differentiable feature selection and
  reconstruction.
\newblock {\em ICML}, 2019.

\bibitem{Armeni2018}
Iro Armeni, Sasha Sax, Amir~Roshan Zamir, and Silvio Savarese.
\newblock Joint {2D-3D}-semantic data for indoor scene understanding.
\newblock {\em CoRR}, abs/1702.01105, 2017.

\bibitem{Atzmon2018}
Matan Atzmon, Haggai Maron, and Yaron Lipman.
\newblock Point convolutional neural networks by extension operators.
\newblock {\em ACM Trans. Graph.}, 37(4), July 2018.

\bibitem{Behley_2019_ICCV}
Jens Behley, Martin Garbade, Andres Milioto, Jan Quenzel, Sven Behnke, Cyrill
  Stachniss, and Jurgen Gall.
\newblock {SemanticKITTI}: A dataset for semantic scene understanding of
  {LiDAR} sequences.
\newblock In {\em The IEEE International Conference on Computer Vision (ICCV)},
  October 2019.

\bibitem{Boulch2017}
Alexandre Boulch, Bertrand~Le Saux, and Nicolas Audebert.
\newblock Unstructured point cloud semantic labeling using deep segmentation
  networks.
\newblock In {\em Proceedings of the Workshop on 3D Object Retrieval}, 3Dor
  ’17, page 17–24, Goslar, DEU, 2017. Eurographics Association.

\bibitem{Brock2016}
Andr{\'{e}} Brock, Theodore Lim, James~M. Ritchie, and Nick Weston.
\newblock Generative and discriminative voxel modeling with convolutional
  neural networks.
\newblock {\em CoRR}, abs/1608.04236, 2016.

\bibitem{chen2019lsanet}
Lin-Zhuo Chen, Xuan-Yi Li, Deng-Ping Fan, Ming-Ming Cheng, Kai Wang, and
  Shao-Ping Lu.
\newblock {LSANet}: Feature learning on point sets by local spatial aware
  layer.
\newblock {\em arXiv preprint arXiv:1905.05442}, 2019.

\bibitem{Chen_2017_CVPR}
Xiaozhi Chen, Huimin Ma, Ji Wan, Bo Li, and Tian Xia.
\newblock Multi-view {3D} object detection network for autonomous driving.
\newblock In {\em The IEEE Conference on Computer Vision and Pattern
  Recognition (CVPR)}, July 2017.

\bibitem{clevert2015fast}
Djork\-Arn{\'e} Clevert, Thomas Unterthiner, and Sepp Hochreiter.
\newblock Fast and accurate deep network learning by exponential linear units
  (elus).
\newblock {\em arXiv preprint arXiv:1511.07289}, 2015.

\bibitem{NIPS2019_8962}
Theo Deprelle, Thibault Groueix, Matthew Fisher, Vladimir Kim, Bryan Russell,
  and Mathieu Aubry.
\newblock Learning elementary structures for {3D} shape generation and
  matching.
\newblock In H. Wallach, H. Larochelle, A. Beygelzimer, F. d\textquotesingle
  Alch\'{e}-Buc, E. Fox, and R. Garnett, editors, {\em Advances in Neural
  Information Processing Systems 32}, pages 7435--7445. Curran Associates,
  Inc., 2019.

\bibitem{devlin2018bert}
Jacob Devlin, Ming-Wei Chang, Kenton Lee, and Kristina Toutanova.
\newblock {BERT}: Pre-training of deep bidirectional transformers for language
  understanding.
\newblock {\em arXiv preprint arXiv:1810.04805}, 2018.

\bibitem{dixon1987mathographics}
Robert Dixon.
\newblock Mathographics. basic blackwell limited, 1987.

\bibitem{Dovrat_2019_CVPR}
Oren Dovrat, Itai Lang, and Shai Avidan.
\newblock Learning to sample.
\newblock In {\em The IEEE Conference on Computer Vision and Pattern
  Recognition (CVPR)}, June 2019.

\bibitem{GroWieLen18}
F. Groh*, P. Wieschollek*, and H.~P.~A. Lensch.
\newblock Flex-convolution (million-scale point-cloud learning beyond
  grid-worlds).
\newblock In {\em Computer Vision - ACCV 2018 - 14th Asian Conference on
  Computer Vision}, Dec. 2018.
\newblock *equal contribution.

\bibitem{Groueix_2018_ECCV}
Thibault Groueix, Matthew Fisher, Vladimir~G. Kim, Bryan~C. Russell, and
  Mathieu Aubry.
\newblock {3D-CODED}: {3D} correspondences by deep deformation.
\newblock In {\em The European Conference on Computer Vision (ECCV)}, September
  2018.

\bibitem{hackel2017isprs}
Timo Hackel, N. Savinov, L. Ladicky, Jan~D. Wegner, K. Schindler, and M.
  Pollefeys.
\newblock {SEMANTIC3D.NET}: A new large-scale point cloud classification
  benchmark.
\newblock In {\em ISPRS Annals of the Photogrammetry, Remote Sensing and
  Spatial Information Sciences}, volume IV-1-W1, pages 91--98, 2017.

\bibitem{NIPS2017_6703}
Will Hamilton, Zhitao Ying, and Jure Leskovec.
\newblock Inductive representation learning on large graphs.
\newblock In I. Guyon, U.~V. Luxburg, S. Bengio, H. Wallach, R. Fergus, S.
  Vishwanathan, and R. Garnett, editors, {\em Advances in Neural Information
  Processing Systems 30}, pages 1024--1034. Curran Associates, Inc., 2017.

\bibitem{hu2019randla}
Qingyong Hu, Bo Yang, Linhai Xie, Stefano Rosa, Yulan Guo, Zhihua Wang, Niki
  Trigoni, and Andrew Markham.
\newblock {RandLA-Net}: Efficient semantic segmentation of large-scale point
  clouds.
\newblock {\em Proceedings of the IEEE Conference on Computer Vision and
  Pattern Recognition}, 2020.

\bibitem{Hua_2018_CVPR}
Binh-Son Hua, Minh-Khoi Tran, and Sai-Kit Yeung.
\newblock Pointwise convolutional neural networks.
\newblock In {\em The IEEE Conference on Computer Vision and Pattern
  Recognition (CVPR)}, June 2018.

\bibitem{Huang_2018_CVPR}
Qiangui Huang, Weiyue Wang, and Ulrich Neumann.
\newblock Recurrent slice networks for {3D} segmentation of point clouds.
\newblock In {\em The IEEE Conference on Computer Vision and Pattern
  Recognition (CVPR)}, June 2018.

\bibitem{Jiang_2019_ICCV}
Li Jiang, Hengshuang Zhao, Shu Liu, Xiaoyong Shen, Chi-Wing Fu, and Jiaya Jia.
\newblock Hierarchical point-edge interaction network for point cloud semantic
  segmentation.
\newblock In {\em The IEEE International Conference on Computer Vision (ICCV)},
  October 2019.

\bibitem{Klokov_2017_ICCV}
Roman Klokov and Victor Lempitsky.
\newblock Escape from cells: Deep kd-networks for the recognition of {3D} point
  cloud models.
\newblock In {\em The IEEE International Conference on Computer Vision (ICCV)},
  Oct 2017.

\bibitem{Komarichev_2019_CVPR}
Artem Komarichev, Zichun Zhong, and Jing Hua.
\newblock {A-CNN}: Annularly convolutional neural networks on point clouds.
\newblock In {\em The IEEE Conference on Computer Vision and Pattern
  Recognition (CVPR)}, June 2019.

\bibitem{Lan_2019_CVPR}
Shiyi Lan, Ruichi Yu, Gang Yu, and Larry~S. Davis.
\newblock Modeling local geometric structure of {3D} point clouds using
  {Geo-CNN}.
\newblock In {\em The IEEE Conference on Computer Vision and Pattern
  Recognition (CVPR)}, June 2019.

\bibitem{Landrieu_2018_CVPR}
Loic Landrieu and Martin Simonovsky.
\newblock Large-scale point cloud semantic segmentation with superpoint graphs.
\newblock In {\em The IEEE Conference on Computer Vision and Pattern
  Recognition (CVPR)}, June 2018.

\bibitem{Lei_2019_CVPR}
Huan Lei, Naveed Akhtar, and Ajmal Mian.
\newblock Octree guided {CNN} with spherical kernels for {3D} point clouds.
\newblock In {\em The IEEE Conference on Computer Vision and Pattern
  Recognition (CVPR)}, June 2019.

\bibitem{Li_2018_CVPR}
Jiaxin Li, Ben~M. Chen, and Gim Hee~Lee.
\newblock {SO-Net}: Self-organizing network for point cloud analysis.
\newblock In {\em The IEEE Conference on Computer Vision and Pattern
  Recognition (CVPR)}, June 2018.

\bibitem{NIPS2018_7362}
Yangyan Li, Rui Bu, Mingchao Sun, Wei Wu, Xinhan Di, and Baoquan Chen.
\newblock {PointCNN}: Convolution on $\mathcal{X}$-transformed points.
\newblock In S. Bengio, H. Wallach, H. Larochelle, K. Grauman, N. Cesa-Bianchi,
  and R. Garnett, editors, {\em Advances in Neural Information Processing
  Systems 31}, pages 820--830. Curran Associates, Inc., 2018.

\bibitem{Loshchilov2016}
Ilya Loshchilov and Frank Hutter.
\newblock {SGDR:} stochastic gradient descent with restarts.
\newblock {\em CoRR}, abs/1608.03983, 2016.

\bibitem{Mao_2019_ICCV}
Jiageng Mao, Xiaogang Wang, and Hongsheng Li.
\newblock Interpolated convolutional networks for {3D} point cloud
  understanding.
\newblock In {\em The IEEE International Conference on Computer Vision (ICCV)},
  October 2019.

\bibitem{martins2016softmax}
Andre Martins and Ramon Astudillo.
\newblock From softmax to sparsemax: A sparse model of attention and
  multi-label classification.
\newblock In {\em International Conference on Machine Learning}, pages
  1614--1623, 2016.

\bibitem{Maturana2015}
Daniel {Maturana} and Sebastian {Scherer}.
\newblock {VoxNet}: A {3D} convolutional neural network for real-time object
  recognition.
\newblock In {\em 2015 IEEE/RSJ International Conference on Intelligent Robots
  and Systems (IROS)}, pages 922--928, 2015.

\bibitem{Milioto2019}
Andres {Milioto}, Ignacio {Vizzo}, Jens {Behley}, and Cyrill {Stachniss}.
\newblock {RangeNet} ++: Fast and accurate {LiDAR} semantic segmentation.
\newblock In {\em 2019 IEEE/RSJ International Conference on Intelligent Robots
  and Systems (IROS)}, pages 4213--4220, 2019.

\bibitem{Monti_2017_CVPR}
Federico Monti, Davide Boscaini, Jonathan Masci, Emanuele Rodola, Jan Svoboda,
  and Michael~M. Bronstein.
\newblock Geometric deep learning on graphs and manifolds using mixture model
  {CNNs}.
\newblock In {\em The IEEE Conference on Computer Vision and Pattern
  Recognition (CVPR)}, July 2017.

\bibitem{Qi_2018_CVPR}
Charles~R. Qi, Wei Liu, Chenxia Wu, Hao Su, and Leonidas~J. Guibas.
\newblock Frustum pointnets for {3D} object detection from {RGB-D} data.
\newblock In {\em The IEEE Conference on Computer Vision and Pattern
  Recognition (CVPR)}, June 2018.

\bibitem{Qi_2017_CVPR}
Charles~R. Qi, Hao Su, Kaichun Mo, and Leonidas~J. Guibas.
\newblock {PointNet}: Deep learning on point sets for {3D} classification and
  segmentation.
\newblock In {\em The IEEE Conference on Computer Vision and Pattern
  Recognition (CVPR)}, July 2017.

\bibitem{Qi_2016_CVPR}
Charles~R. Qi, Hao Su, Matthias Niessner, Angela Dai, Mengyuan Yan, and
  Leonidas~J. Guibas.
\newblock Volumetric and multi-view cnns for object classification on {3D}
  data.
\newblock In {\em The IEEE Conference on Computer Vision and Pattern
  Recognition (CVPR)}, June 2016.

\bibitem{NIPS2017_7095}
Charles~Ruizhongtai Qi, Li Yi, Hao Su, and Leonidas~J Guibas.
\newblock {PointNet++}: Deep hierarchical feature learning on point sets in a
  metric space.
\newblock In I. Guyon, U.~V. Luxburg, S. Bengio, H. Wallach, R. Fergus, S.
  Vishwanathan, and R. Garnett, editors, {\em Advances in Neural Information
  Processing Systems 30}, pages 5099--5108. Curran Associates, Inc., 2017.

\bibitem{Roynard2018}
Xavier Roynard, Jean{-}Emmanuel Deschaud, and Fran{\c{c}}ois Goulette.
\newblock Classification of point cloud scenes with multiscale voxel deep
  network.
\newblock {\em CoRR}, abs/1804.03583, 2018.

\bibitem{Simonovsky_2017_CVPR}
Martin Simonovsky and Nikos Komodakis.
\newblock Dynamic edge-conditioned filters in convolutional neural networks on
  graphs.
\newblock In {\em The IEEE Conference on Computer Vision and Pattern
  Recognition (CVPR)}, July 2017.

\bibitem{Su_2018_CVPR}
Hang Su, Varun Jampani, Deqing Sun, Subhransu Maji, Evangelos Kalogerakis,
  Ming-Hsuan Yang, and Jan Kautz.
\newblock {SPLATNet}: Sparse lattice networks for point cloud processing.
\newblock In {\em The IEEE Conference on Computer Vision and Pattern
  Recognition (CVPR)}, June 2018.

\bibitem{Tatarchenko_2018_CVPR}
Maxim Tatarchenko, Jaesik Park, Vladlen Koltun, and Qian-Yi Zhou.
\newblock Tangent convolutions for dense prediction in {3D}.
\newblock In {\em The IEEE Conference on Computer Vision and Pattern
  Recognition (CVPR)}, June 2018.

\bibitem{Tchapmi2017}
Lyne {Tchapmi}, Christopher {Choy}, Iro {Armeni}, JunYoung {Gwak}, and Silvio
  {Savarese}.
\newblock {SEGCloud}: Semantic segmentation of {3D} point clouds.
\newblock In {\em 2017 International Conference on 3D Vision (3DV)}, pages
  537--547, 2017.

\bibitem{Thomas2018}
Hugues {Thomas}, François {Goulette}, Jean-Emmanuel {Deschaud}, Beatriz
  {Marcotegui}, and Yann {LeGall}.
\newblock Semantic classification of {3D} point clouds with multiscale
  spherical neighborhoods.
\newblock In {\em 2018 International Conference on 3D Vision (3DV)}, pages
  390--398, 2018.

\bibitem{Thomas_2019_ICCV}
Hugues Thomas, Charles~R. Qi, Jean-Emmanuel Deschaud, Beatriz Marcotegui,
  Francois Goulette, and Leonidas~J. Guibas.
\newblock {KPConv}: Flexible and deformable convolution for point clouds.
\newblock In {\em The IEEE International Conference on Computer Vision (ICCV)},
  October 2019.

\bibitem{NIPS2017_7181}
Ashish Vaswani, Noam Shazeer, Niki Parmar, Jakob Uszkoreit, Llion Jones,
  Aidan~N Gomez, \L~ukasz Kaiser, and Illia Polosukhin.
\newblock Attention is all you need.
\newblock In I. Guyon, U.~V. Luxburg, S. Bengio, H. Wallach, R. Fergus, S.
  Vishwanathan, and R. Garnett, editors, {\em Advances in Neural Information
  Processing Systems 30}, pages 5998--6008. Curran Associates, Inc., 2017.

\bibitem{velickovic2018graph}
Petar Veli{\v{c}}kovi{\'{c}}, Guillem Cucurull, Arantxa Casanova, Adriana
  Romero, Pietro Li{\`{o}}, and Yoshua Bengio.
\newblock {Graph Attention Networks}.
\newblock {\em International Conference on Learning Representations}, 2018.
\newblock accepted as poster.

\bibitem{Verma_2018_CVPR}
Nitika Verma, Edmond Boyer, and Jakob Verbeek.
\newblock {FeaStNet}: Feature-steered graph convolutions for {3D} shape
  analysis.
\newblock In {\em The IEEE Conference on Computer Vision and Pattern
  Recognition (CVPR)}, June 2018.

\bibitem{Wang_2018_ECCV}
Chu Wang, Babak Samari, and Kaleem Siddiqi.
\newblock Local spectral graph convolution for point set feature learning.
\newblock In {\em The European Conference on Computer Vision (ECCV)}, September
  2018.

\bibitem{Wang_2019_CVPR}
Lei Wang, Yuchun Huang, Yaolin Hou, Shenman Zhang, and Jie Shan.
\newblock Graph attention convolution for point cloud semantic segmentation.
\newblock In {\em The IEEE Conference on Computer Vision and Pattern
  Recognition (CVPR)}, June 2019.

\bibitem{Wang_2018_CVPR}
Shenlong Wang, Simon Suo, Wei-Chiu Ma, Andrei Pokrovsky, and Raquel Urtasun.
\newblock Deep parametric continuous convolutional neural networks.
\newblock In {\em The IEEE Conference on Computer Vision and Pattern
  Recognition (CVPR)}, June 2018.

\bibitem{Yue_2019}
Yue Wang, Yongbin Sun, Ziwei Liu, Sanjay~E. Sarma, Michael~M. Bronstein, and
  Justin~M. Solomon.
\newblock Dynamic graph {CNN} for learning on point clouds.
\newblock {\em ACM Trans. Graph.}, 38(5), Oct. 2019.

\bibitem{Wu2018}
Bichen {Wu}, Alvin {Wan}, Xiangyu {Yue}, and Kurt {Keutzer}.
\newblock {SqueezeSeg}: Convolutional neural nets with recurrent {CRF} for
  real-time road-object segmentation from {3D} {LiDAR} point cloud.
\newblock In {\em 2018 IEEE International Conference on Robotics and Automation
  (ICRA)}, pages 1887--1893, 2018.

\bibitem{Wu2019}
Bichen {Wu}, Xuanyu {Zhou}, Sicheng {Zhao}, Xiangyu {Yue}, and Kurt {Keutzer}.
\newblock {SqueezeSegV2}: Improved model structure and unsupervised domain
  adaptation for road-object segmentation from a {LiDAR} point cloud.
\newblock In {\em 2019 International Conference on Robotics and Automation
  (ICRA)}, pages 4376--4382, 2019.

\bibitem{Wu_2019_CVPR}
Wenxuan Wu, Zhongang Qi, and Li Fuxin.
\newblock Pointconv: Deep convolutional networks on 3d point clouds.
\newblock In {\em The IEEE Conference on Computer Vision and Pattern
  Recognition (CVPR)}, June 2019.

\bibitem{Wu_2015_CVPR}
Zhirong Wu, Shuran Song, Aditya Khosla, Fisher Yu, Linguang Zhang, Xiaoou Tang,
  and Jianxiong Xiao.
\newblock {3D ShapeNets}: A deep representation for volumetric shapes.
\newblock In {\em The IEEE Conference on Computer Vision and Pattern
  Recognition (CVPR)}, June 2015.

\bibitem{Ye_2018_ECCV}
Xiaoqing Ye, Jiamao Li, Hexiao Huang, Liang Du, and Xiaolin Zhang.
\newblock {3D} recurrent neural networks with context fusion for point cloud
  semantic segmentation.
\newblock In {\em The European Conference on Computer Vision (ECCV)}, September
  2018.

\bibitem{Zhang_2019_ICCV}
Zhiyuan Zhang, Binh-Son Hua, and Sai-Kit Yeung.
\newblock {ShellNet}: Efficient point cloud convolutional neural networks using
  concentric shells statistics.
\newblock In {\em The IEEE International Conference on Computer Vision (ICCV)},
  October 2019.

\bibitem{zhao2019pointweb}
Hengshuang Zhao, Li Jiang, Chi-Wing Fu, and Jiaya Jia.
\newblock {PointWeb}: Enhancing local neighborhood features for point cloud
  processing.
\newblock In {\em CVPR}, 2019.

\bibitem{Zhao_2019_CVPR}
Hengshuang Zhao, Li Jiang, Chi-Wing Fu, and Jiaya Jia.
\newblock {PointWeb}: Enhancing local neighborhood features for point cloud
  processing.
\newblock In {\em The IEEE Conference on Computer Vision and Pattern
  Recognition (CVPR)}, June 2019.

\bibitem{Zhou_2018_CVPR}
Yin Zhou and Oncel Tuzel.
\newblock {VoxelNet}: End-to-end learning for point cloud based {3D} object
  detection.
\newblock In {\em The IEEE Conference on Computer Vision and Pattern
  Recognition (CVPR)}, June 2018.

\end{thebibliography}
\bibliographystyle{ieee_fullname}
%
%
%
%

\end{document}